\documentclass[letterpaper, 10 pt, conference]{ieeeconf}
\IEEEoverridecommandlockouts
\usepackage{amsmath,amssymb,amsfonts}
\usepackage{algorithmic}
\usepackage{graphicx}
\usepackage{textcomp}
\usepackage{xcolor}
\def\BibTeX{{\rm B\kern-.05em{\sc i\kern-.025em b}\kern-.08em
    T\kern-.1667em\lower.7ex\hbox{E}\kern-.125emX}}

\usepackage{etoolbox} 
\usepackage{hyperref}

\usepackage{multicol}
\usepackage{multirow}
\usepackage{booktabs}
\usepackage{makecell}

\usepackage{siunitx}
\usepackage{cleveref}
\usepackage{todonotes}

\usepackage{listings}
\usepackage{xcolor}

\usepackage[ruled,vlined]{algorithm2e}

\usepackage[style=ieee,citestyle=numeric-comp,sorting=none]{biblatex}
\title{\LARGE \bf
Route Fragmentation Based on Resource-centric Prioritisation for Efficient Multi-Robot Path Planning in Agricultural Environments
}

\author{
    James R. Heselden$^{1,2,3}$, and Gautham P. Das$^{1,2,4,*}$
    \thanks{
        $^{1}$Lincoln Centre for Autonomous Systems, University of Lincoln, UK. $^{2}$Lincoln Institute for Agri-food Technology, University of Lincoln, UK. $^{3}$ORCiD: James R. Heselden (0000-0001-6494-4981). $^{4}$ORCiD: Gautham P. Das (0000-0001-5351-9533) $^{*}$Corresponding Author (gdas@lincoln.ac.uk).
    }
}

\begin{document}
\maketitle


\thispagestyle{empty}
\pagestyle{empty}

\begin{abstract}

Agricultural environments present high proportions of spatially dense navigation bottlenecks for long-term navigation and operational planning of agricultural mobile robots. The existing agent-centric multi-robot path planning (MRPP) approaches resolve conflicts from the perspective of agents, rather than from the resources under contention. Further, the density of such contentions limits the capabilities of spatial interleaving, a concept that many planners rely on to achieve high throughput. In this work, two variants of the priority-based Fragment Planner (FP) are presented as resource-centric MRPP algorithms that leverage route fragmentation to enable partial route progression and limit the impact of binary-based waiting. These approaches are evaluated in lifelong simulation over a 3.6km topological map representing a commercial polytunnel environment. Their performances are contrasted against 5 baseline algorithms with varying robotic fleet sizes. The Fragment Planners achieved significant gains in throughput compared with Prioritised Planning (PP) and Priority-Based Search (PBS) algorithms. They further demonstrated a task throughput of 95\% of the optimal task throughput over the same time period. This work shows that, for long-term deployment of agricultural robots in corridor-dominant agricultural environments, resource-centric MRPP approaches are a necessity for high-efficacy operational planning.

\end{abstract}


\section{Introduction}\label{sec:int}

Lifelong MRPP remains challenging for long-term autonomous deployments. Much of the work in this domain focuses on well-structured warehouse environments or free-space navigation. Agricultural environments, present a less well-explored set of environmental arrangements and in turn unique coordination challenges as a result of dense spatial bottlenecks that limit agent-centric path planning approaches. Handling these bottlenecks is a core challenge in MRPP approaches as their impact within agent-centric planning increases as fleet sizes scales \cite{Prorok2021}. To address this, many agricultural MRS deployments utilise small fleet sizes with spatial segmentation in order to guarantee fleet safety. Whilst this is reliable for small fleets, the lack of shared operational space prevents systems from achieving full potential and results in cost-inefficiencies. To improve efficiency and throughput, it is critical that robots are able to both share their operational environment and scale to larger fleet sizes.


It is well-established in the literature that coupled MRPP approaches suffer from scaling due to exponential state space exploration \cite{LaValle1999}. As such, decoupled MRPP approaches are more widely adopted in practice. However, many approaches assume full spatial flexibility, open grids, and alternate-path availability \cite{Wen2022a, Andreychuk2019a}.
Many decoupled approaches operate with the use of predetermined forced global agent-centric orders. Prioritised Planning (PP) \cite{Heselden2021} is a clear example of this, utilising fixed robot ordering to prevent conflicts in the planning stage, and as such, operating in linear time. Priority-Based Search (PBS) is an approach which loosens the enforcement of predetermined global orders, instead building partial orders dynamically over space-time expanded contention and performing repeat replans to couple priority orderings to contention resolution \cite{Ma2019}.

The PBS approach, despite loosening the enforcement of fixed global orders, still relies on agent-centric priorities. Agent-centric orders are reliable when collisions are sparse and interleaving is available. However, in agricultural environments, interleaving is not always a viable option. Agricultural environments, e.g., polytunnels and open-fields, often have strict navigational constraints more closely modelling corridor-like topologies. These prevent temporal interleaving and worsen agent-centric approaches as the collisions are often bottlenecks which are being resolved at the agent-level, rather than locally.

To ensure planning is reliable without coupled planning in these environmental arrangements, it is thus necessary for planners to focus on resource-centric priority ownership rather than agent-centric orderings. To this end, this work presents two variants (space-only-aware and space-time-aware) of the Fragment Planner (FP), a decoupled approach which uses priority-orderings over resources to resolve contention with a single pass. The approach further utilises fragment-level route execution to enable continuous route execution towards contested bottlenecks rather than utilising binary-based waiting for full-path availability.

\section{Problem Definition}\label{sec:lit}

\subsection{Topological Representation}
To effectively manage a fleet of robots in real time, it is important that the system is able to scale with the number of robots and with the size of the environment. As the joint state space increases exponentially with this complexity, it is important to be able to discretise the environment in a way that minimises the state space without reducing the operation of the fleet. In this work, a topological map is utilised to represent the traversable paths through the environment.

Presented in \cite{das2023unified}, the topological map acts as a unified representation of the environment as a network graph in which nodes act as potential crossing points and are connected with edges. Formally, let the environment be represented as a directed graph $G = (V, E)$, where $V$ denotes the set of traversable nodes and $E \subseteq V \times V$ denotes the set of directed edges. These edges encode traversability constraints of the environment, such as maximum allowed speed, navigation parameters, terrain type, and permissible envelope size. Centralised MRPP is conducted within this topological space in order to reduce the joint state space and avoid the complexity of metric spaces.

In planning, routes are generated as an ordered sequence of nodes, where for each agent $a \in \mathcal{A}$ a route is defined as $R_a = \langle v_0, v_1, \dots, v_k \rangle$ with $v_i \in V$ and $(v_i, v_{i+1}) \in E$ for all $i$. The primary aim of route planning being to avoid points of contention wherein a node $v \in V$ is defined as being in contention if $\exists\, a \neq b \in \mathcal{A}$ such that $v \in R_a \cap R_b$.

Unlike a basic network representation, incorporating these traversability constraints within the map itself enables the map to stay coupled with the environment it represents. This also allows planners to utilise these parameters in their optimisation process. To this end, in this work, the robot's footprint is utilised along with the topological map's permissible envelope to create permissible sub-graphs for robots to plan over and reduce the planning space per robot. Similarly, prioritisation processes are able to utilise per-edge speed limits along with per-robot nominal speeds in planning calculations.

\begin{figure*}[t!]
    \centering
    \includegraphics[width=.42\linewidth]{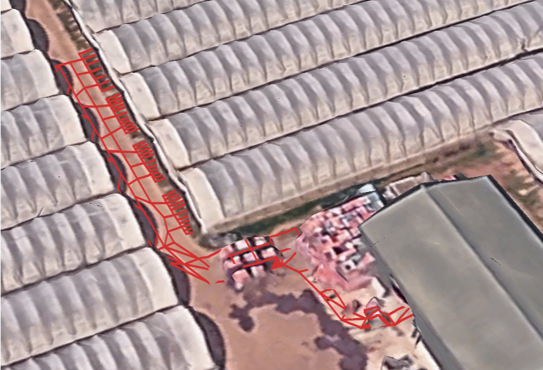}
    \includegraphics[width=.42\linewidth]{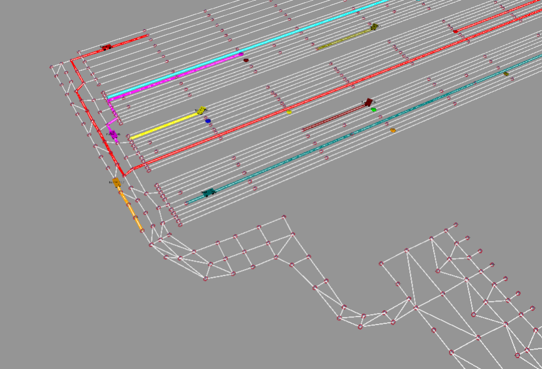}
    \caption{Polytunnel environment presented with an aerial photogrammetry render (left), and the topological representation from simulation (right).}
    \label{fig:environment}
\end{figure*}

In this work, we consider an agricultural polytunnel environment with raised tables for the plants. The narrow rows between the raised tables act as narrow and long corridors traversible by robots. The robots are unable to move under the raised beds due to their size as well as the foliage of the plants on the raised beds. Under the topological representation, this models the environment as a series of end-connected narrow corridors as shown in Figure \ref{fig:environment}. 

\subsection{Corridor-Dominated Topology}

This type of environment, characterised by a high proportion of nodes satisfying $\deg(v) \leq 2$, results in long linked paths connected by infrequent branching points. Further, they contain high proportions of articulation nodes, where the removal of a single node $v \in V$ can result in large sections of the map becoming inaccessible.

It is topologically distinct from environments constructed of open spaces or connected rooms \cite{Ryan2008}. The topological structure results in unique planning requirements with heavy constraints on overtaking and interleaving. Under this structure, alternative paths are limited and contention is spatially concentrated at bottleneck nodes. In turn, this increases the impact of local resource contention during coordination.

\subsection{Priority-Based Planning}
Within MRPP, there are two main approaches to planning routes, coupled and decoupled path planning \cite{heselden2023heuristics}. Within coupled path planning, all robots in a single route search instance are planned with direct consideration of one another. In contrast, decoupled path planning generates each robot's path independently, and then goes through a conflict resolution stage that identifies nodes or fragments of the routes that overlap and implements mitigating policies. Due to the exponential increase in the joint state as fleet size increases, coupled path planners often respond poorly to scalability.

Decoupled path planners, perform well under increasing fleet sizes. However, they are often unable to guarantee a global optimum plan. The most prominent decoupled path planning approach is prioritised planning \cite{Erdmann1987}. In this approach, a pre-selected heuristic is used to provide a global ordering $\prec$ over agents in $\mathcal{A}$, wherein agents are planned sequentially such that each agent $a$ generates its route while treating the routes of all agents $b \prec a$ as fixed occupancy constraints. This approach is able to provide good scaling to increases in fleet size as planning is completed in linear time.

The heuristic chosen can be simple prioritisation based on name \cite{Clark2001} or path length \cite{VanDenBerg2005}, or can be more complicated such as utilising the homology classes of trajectories \cite{Wu2019}. There are many such priority-focused examples of decoupled MRPP as outlined in \cite{heselden2023heuristics}. In this work, the Fragment Planner, is presented as a novel priority-based planner which focuses on resource-centric prioritisation to resolve local ownership of critical points.
 
\section{Methodology}\label{sec:meth}

A modular dynamic multi-robot path planning and navigation execution framework is developed as a single framework on which different MRPP algorithms with standardised interfaces can be easily tested. To ensure planners are all compatible with the core deployment architecture, a BasePlanner is developed with a shared, consistent behaviour. The shared functionalities available for all planners are outlined in Section. \ref{ssec:base_planner}. Five agent-centric prioritisation approaches are explored in this paper - A greedy-search-based naive decoupled Naïve planner, three generic prioritised planners, and the PBS planner. These are discussed in Sections. \ref{ssec:naive_planner}, \ref{ssec:prioritised_planners} and \ref{ssec:pbs_planner} respectively. The fragment planners (Space-Only-Aware and Space-Time-Aware), which prioiritise the agents based on a resouce-centric prioiritisation approach, are described in Section \ref{ssec:fragment_planners}. Table \ref{tab:planner_comparison} summarises the conceptual differences between the approaches evaluated.

\begin{table*}[ht]
    \centering
    \caption{Conceptual differences in evaluated approaches}
    \label{tab:planner_comparison}
    \renewcommand{\arraystretch}{1.25}
    \begin{tabular}{
        |>{\centering\arraybackslash}p{2.6cm}
        |>{\centering\arraybackslash}p{2.0cm}
        |>{\centering\arraybackslash}p{2.3cm}
        |>{\centering\arraybackslash}p{1.8cm}
        |>{\centering\arraybackslash}p{1.4cm}
        |>{\centering\arraybackslash}p{2.1cm}
        |>{\centering\arraybackslash}p{2.0cm}|
    }
        \hline
        \textbf{Approach} &
        \textbf{Conflict Resolution Trigger} &
        \textbf{Temporal Modelling} &
        \textbf{Temporal Priority Ordering} &
        \textbf{Enforced Priority Ordering} &
        \textbf{Priority Ordering Scope} &
        \textbf{Conflict Responses} \\
        \hline
        \textbf{Naïve Independent Planning} &
        N/A &
        N/A &
        N/A &
        N/A &
        N/A &
        Waiting Only \\
        \hline
        \textbf{Traditional Prioritised Planner} &
        N/A &
        N/A &
        Global &
        Forced &
        Agent-Level &
        Waiting Only \\
        \hline
        \textbf{Priority Based Search (using local priority adjustment)} &
        During Planning, on Conflict &
        Space--Time Expanded (Future Conflicts Found) &
        Global &
        Forced &
        Agent-Centric &
        Priority Constraint Search \\
        \hline
        \textbf{Fragment Planner (Space-Only-Aware)} &
        During Planning, on Contention &
        Timeless &
        Spatially Local Fragment Ownership &
        Forced (Local) &
        Spatially Focused Resource-Centric &
        Waiting / Partial Route Execution / Spatial Deferral \\
        \hline
        \textbf{Fragment Planner (Space-Time-Aware)} &
        During Planning, on Contention &
        Time-Aware &
        Spatially Local Fragment Ownership &
        Forced (Local) &
        Spatially Focused Resource-Centric &
        Waiting / Partial Route Execution / Spatial Deferral \\
        \hline
    \end{tabular}
    \vspace{-1em}
\end{table*}


\subsection{BasePlanner - Shared Implementation Functionalities}
\label{ssec:base_planner}

\subsubsection{Group Assignments}
In order to effectively manage lifelong framing, it is important that the planning state is able to be carried over between planning instances. To that end, we utilise a group system wherein all agents in the agent set $\mathcal{A}$ are categorised into one of three groups: New Active ($\mathcal{N}$, has not received a plan before), Active ($\mathcal{C}$, is executing a plan), and Inactive ($\mathcal{I}$, is stationary). These groups are applied according to Algorithm \ref{alg:lifecycle_grouping}, utilising information from the agent's current goal and route taken during the preparation to the start of planning.

\begin{algorithm}[htb]
\caption{BasePlanner: \textsc{GetGroups}}
\label{alg:lifecycle_grouping}
\KwIn{Agent set $\mathcal{A}$, current planning context}
\KwOut{$(\mathcal{N},\mathcal{C},\mathcal{I})$: new, active, inactive groups}
\BlankLine
$\mathcal{N}\leftarrow\emptyset,\ \mathcal{C}\leftarrow\emptyset,\ \mathcal{I}\leftarrow\emptyset$\;
\ForEach{$a\in\mathcal{A}$}{
  \uIf{\textsc{NewlyEligible}$(a)$}{$\mathcal{N}\leftarrow\mathcal{N}\cup\{a\}$}
  \uElseIf{\textsc{Active}$(a)$}{$\mathcal{C}\leftarrow\mathcal{C}\cup\{a\}$}
  \Else{$\mathcal{I}\leftarrow\mathcal{I}\cup\{a\}$}
}
\Return $(\mathcal{N},\mathcal{C},\mathcal{I})$\;
\end{algorithm}

\subsubsection{Route for Active}
For planning routes for active agents, all planners utilise the same basic workflow in order to function within the lifelong coordinated space. Algorithm \ref{alg:base_route} outlines three stages to this process, wherein a pre-routing configuration is performed to prepare the agent for planning, removing any existing route properties. Then its endpoints (start node $s$ and goal node $g$) are identified, and the map is filtered using the provided agent set $\mathcal{A}$ and blocking set $\mathcal{B}$. The pre-routing validation is then performed to verify if a routing from $s$ to $g$ is possible (see Algorithm \ref{alg:shared_checks}) and only then is the route search processed. The final generated route is then post-processed to move the new route onto the robot for execution.

\begin{algorithm}[htb]
\caption{BasePlanner: \textsc{RouteForActive}}
\label{alg:base_route}
\KwIn{Agent $a$, agent set $\mathcal{A}$, routing map $map$, blocking set $\mathcal{B}$}
\KwOut{Updated route state for $a$ (or inactive)}
\BlankLine
\textsc{PreRoutingConfigure}(a)\;
$(s,g)\leftarrow\textsc{GetRouteEndpoints}(a)$\;
\If{$map=\texttt{filtered}$}{
  $\textsc{GenerateFilteredMap}(a,s,g,\mathcal{A},\mathcal{B})$\;
}
\If{$\neg\textsc{AllowRouting}(a,s,g)$}{
  \Return $(\texttt{inactive},a)$\;
}
$r\leftarrow \textsc{RouteSearch}(s,g)$\;
\Return \textsc{PostRoutingConfigure}(a,r)\;
\end{algorithm}

\subsubsection{Routing Checks}
To ensure a routing is viable, the start and goal nodes for are validated according to Algorithm \ref{alg:shared_checks}. This algorithm outlines three key checks to verify that the start and goal nodes exist and are within the map, and that planning is required at this stage. If any stage fails, the route search will move the agent from active to inactive for separate route processing.

\begin{algorithm}[t]
\caption{BasePlanner: \textsc{AllowRouting}}
\label{alg:shared_checks}
\KwIn{Agent $a$, start $s$, goal $g$}
\KwOut{Confirmation on if routing is viable}
\BlankLine
\If{$g=\varnothing$}{
  \Return $\textsc{False}$\;
}
\If{$s=g$}{
  \Return $\textsc{False}$\;
}
\If{$\neg\textsc{IsValidNode}(s)$ \textbf{or} $\neg\textsc{IsValidNode}(g)$}{
  \Return $\textsc{False}$\;
}
\Return \textsc{True}\;
\end{algorithm}

\begin{algorithm}[h]
\caption{BasePlanner: \textsc{GetPriorities}}
\label{alg:base_get_orderings}
\KwIn{Groups $\mathcal{N},\mathcal{C},\mathcal{I}$, heuristic $h$, optional target node $v$}
\KwOut{Ordered agent-id list $ord$}
\BlankLine
$\mathcal{U}\leftarrow \mathcal{N}\cup\mathcal{C}\cup\mathcal{I}$,\quad $score\leftarrow\emptyset$\;
$names \leftarrow \textsc{SortedIds}(\mathcal{U})$\;
\ForEach{$a\in\mathcal{U}$}{
  \uIf{$h=\texttt{name}$}{
    $score[a.id]\leftarrow \textsc{Index}(names,a.id)$
  }
  \uElseIf{$h=\texttt{time\_since\_task\_start}$}{
    $score[a.id]\leftarrow a.\textit{task\_start\_time}$
  }
  \uElseIf{$h=\texttt{closest\_first}$}{
    $score[a.id]\leftarrow |a.\textit{optimal\_route}|$
  }
  \uElseIf{$h=\texttt{distance\_to\_node}$}{
    $\ell_a \leftarrow \textsc{RoutePrefixDistance}(a, v)$\;
    $score[a.id]\leftarrow \ell_a$
  }
  \uElseIf{$h=\texttt{time\_to\_node}$}{
    $\ell_a \leftarrow \textsc{RoutePrefixDistance}(a, v)$\;
    $u_a \leftarrow \max(\epsilon, a.\textit{nominal\_speed})$\;
    $score[a.id]\leftarrow \ell_a / u_a$
  }
}
$ord \leftarrow \textsc{SortKeysByValueAscending}(score)$\;
\Return $ord$\;
\end{algorithm}

\subsubsection{Priorities Calculations}
To ensure priority calculation is well-refined, the priority generation system is kept in the base class for all planners to have consistent results. Implemented under this algorithm are five prioritisation heuristics which are able to be utilised. Algorithm \ref{alg:base_get_orderings} outlines the five approaches which let the highest priority agent be the one which: (i) is first alphabetically by name, (ii) has spent the longest on task, (iii) is closest to their goal, (iv) is closest to a point of contention (v) can reach the contention earliest. The agents identifiers are returned in the order of their priority.


\subsection{Naïve Planning}
\label{ssec:naive_planner}
The Naïve Planner (NP) is designed as a baseline example of how the system operates when multi-robot operation is not considered. In this approach, outlined in Algorithm \ref{alg:naïve_find_routes}, each robot is identified and classified into its group, then for each agent requiring planning (i.e., $a \in \mathcal{N} \cup \mathcal{C}$), they are planned over an independent map, generating their optimal route to the node with no consideration of other robots occupying space in the environment (i.e., $B=\phi$) in Algorithm \ref{alg:base_route}.

\begin{algorithm}[t]
\caption{NaïvePlanner: \textsc{FindRoutes}}
\label{alg:naïve_find_routes}
\KwIn{Agent set $\mathcal{A}$}
\KwOut{Updated routes for active/new agents}
\BlankLine
$map \leftarrow \texttt{filtered}$\;
$(\mathcal{N},\mathcal{C},\mathcal{I}) \leftarrow \textsc{GetGroups}(\mathcal{A})$\;
$\mathcal{D}\leftarrow\{a.id\mapsto a \mid a\in\mathcal{N}\cup\mathcal{C}\}$\;
\ForEach{$(a.id)\in\mathcal{D}$}{
  \textsc{RouteForActive}($a$, $A$, $map$, $\emptyset$)\;
}
\end{algorithm}

\begin{algorithm}[h]
\caption{PrioritisedPlanner: \textsc{FindRoutes}}
\label{alg:prioritised_find_routes}
\KwIn{Agent set $\mathcal{A}$, heuristic $h$}
\KwOut{Updated routes for eligible agents}
\BlankLine
$map \leftarrow \texttt{filtered}$\;
$(\mathcal{N},\mathcal{C},\mathcal{I}) \leftarrow \textsc{GetGroups}(\mathcal{A})$\;
$ord \leftarrow \textsc{GetPriorities}(\mathcal{N},\mathcal{C},\mathcal{I},h)$\;
$\mathcal{D}\leftarrow\{a.id\mapsto a \mid a\in\mathcal{N}\cup\mathcal{C}\cup\mathcal{I}\}$\;
\ForEach{$a.id\in ord$}{
  $\mathcal{B} \leftarrow \textsc{OccupiedNodes}(\texttt{route})$\;
  \textsc{RouteForActive}($a$, $A$, $map$, $\mathcal{B}$)\;
}
\end{algorithm}

\subsection{Prioritised Planning}
\label{ssec:prioritised_planners}
The prioritised planner (PP) works according to the principles outlined by \textcite{Erdmann1987}. This approach is broken down into two stages and is outlined in Algorithm \ref{alg:prioritised_find_routes}. Agents are assigned a forced global ordering, using priority scores calculated based on the chosen prioritisation heuristic. Agents are then assigned plans, building onto a global occupancy model. In practice, this is achieved through an iterative planning and sub-graph generation process. In traditional PP implementations, agents which are unable to identify a route must stay stationary until a planning instance arises where a path is able to be achieved. Whilst the PP approach is able to scale linearly due to limited coupling between agents, it is not able to always guarantee a solution for all agents, and even when it does, it is unable to guarantee optimality. In this paper, we consider the following prioritisation heuristics: (1) name, (2) shortest route to the goal and (3) time since the current task start.

\subsection{PBS Planning}
\label{ssec:pbs_planner}
A well-explored process for improving prioritised planning is to incorporate rescheduling. This is an additional step which follows on from the search where by some process, priorities are rearranged between the agents so that a more optimal arrangement can be found. PBS Planning \cite{Ma2019} is a locally space-time expanded prioritised planning implementation which leverages this rescheduling methodology. PBS rescheduling, utilises a depth-first search algorithm to build up partial priority orderings. In this algorithm, outlined in Algorithm \ref{alg:pbs_find_routes_core}, planning priorities are initially empty. Agents are planned decoupled from one another in open space, if any agent generated a route containing a conflict, a partial priority ordering (rule) is created between the two agents involved in the form of a search tree with a branch either way as to which robot goes first. The replanning is then completed again with this partial priority ordering put in place. In practice, agents plan in order of any rules, with sub-graphs being generated for each agent excluding nodes which are occupied by the routes of higher-priority agents.

\begin{algorithm}[t]
\caption{PBSPlanner: \textsc{FindRoutes}}
\label{alg:pbs_find_routes_core}
\KwIn{Agent set $\mathcal{A}$, max attempts $T$}
\KwOut{Conflict-minimised route assignment}
\BlankLine
$map \leftarrow \texttt{filtered}$\;
$(\mathcal{N},\mathcal{C},\mathcal{I})\leftarrow\textsc{GetGroups}(\mathcal{A})$\;
$\mathcal{D}\leftarrow\{a.id\mapsto a\mid a\in\mathcal{N}\cup\mathcal{C}\cup\mathcal{I}\}$\;

$prio[a]\leftarrow -1$ for $a\in\mathcal{N}\cup\mathcal{C}$\;
$prio[a]\leftarrow +\infty$ for $a\in\mathcal{I}$\;
$\mathcal{R}\leftarrow[\emptyset]$ \tcp*{candidate rulesets queue}
$\mathcal{R}_{test}\leftarrow\emptyset$\;

\For{$i\leftarrow 1$ \KwTo $T$}{
  \If{$\mathcal{R}=\emptyset$}{\textbf{break}}
  $r^\star\leftarrow\textsc{PopFirst}(\mathcal{R})$\;
  $prio\leftarrow\textsc{GetPriorities}(r^\star, prio)$\;
  $ord\leftarrow\textsc{SortKeysByValueAscending}(prio)$\;

  \ForEach{$a.id\in ord$}{\textsc{PreRoutingConfigure}($a$)}

  $routes\leftarrow\emptyset$\;
  \ForEach{$a.id\in ord$}{
    $B\leftarrow\textsc{GetAgentsToConsider}(a, r^\star)$\;
    $\mathcal{B} \leftarrow \textsc{OccupiedNodes}(\texttt{a.route},B)$\;
    $ret\leftarrow\textsc{RouteForActive}(a, \mathcal{A},$map$,\mathcal{B})$\;
    \If{$ret=\texttt{inactive}$}{\textsc{RouteForInactive}($a, \mathcal{A})$}
    $routes[a.id]\leftarrow a.route$\;
  }

  $fails\leftarrow\{a.id\mid |routes[a.id]|=0 \land prio[a.id]\neq +\infty\}$\;
  $conf\leftarrow\textsc{GetConflicts}(a, \mathcal{D})$\;

  $\mathcal{R}_{new}\leftarrow\textsc{GetRulesets}(conf,r^\star)$\;
  $\mathcal{R}_{new}\leftarrow\textsc{FilterUntested}(\mathcal{R}_{new},\mathcal{R},\mathcal{R}_{test},r^\star)$\;
  $\mathcal{R}_{test}\leftarrow\mathcal{R}_{test}\cup\{r^\star\}$\;
  $\mathcal{R}\leftarrow\mathcal{R}\cup\mathcal{R}_{new}$\;

  \If{$|conf|=0 \land |fails|=0$}{\textbf{break}}
}
\end{algorithm}


\begin{algorithm}[t]
\caption{FragmentPlanner: \textsc{FindRoutes}}
\label{alg:fragment_find_routes}
\KwIn{Agent set $\mathcal{A}$, priority metric $\pi$}
\KwOut{Updated routes, route fragments, and aligned fragment edges}

$\mathcal{B} \leftarrow \textsc{OccupiedNodes}()$\;
$(\mathcal{N},\mathcal{C},\mathcal{I}) \leftarrow \textsc{GetGroups}()$\;
$\mathcal{U} \leftarrow \mathcal{N}\cup\mathcal{C}$\;
map = \texttt{filtered}\;
$R \leftarrow \{\}$\;

\ForEach{$a \in \mathcal{U}$}{
  $ret \leftarrow \textsc{RouteForActive}(a, \mathcal{A}, map, \mathcal{B})$\;
  \If{$ret=\texttt{inactive}$}{
    $\mathcal{I} \leftarrow \mathcal{I}\cup\{a\}$\;
  }
  \Else{
  $R_{a.id} = ret$\;
  }
}

$(CP,\ CP\_Agents) \leftarrow \textsc{GetCriticalPoints}( R, \mathcal{U}-\mathcal{I})$\;
$F \leftarrow \textsc{AssignFragments}(CP,\ CP\_Agents,\ \pi)$\;
\textsc{BuildRouteFragments}(F)\;
\end{algorithm}

\begin{algorithm}[h]
\caption{FragmentPlanner: \textsc{GetCriticalPoints}}
\label{alg:fragment_get_critical_points}
\KwIn{Agent routes $\{R_a\}$, active-agent set $\mathcal{A}_{goal}$}
\KwOut{$CP,\ CP\_Agents$}
\BlankLine
$CP \leftarrow \emptyset,\ CP\_Agents \leftarrow \emptyset$\;
\ForEach{agent $a$}{
  $CP[\textsc{Key}(R_a)] \leftarrow \emptyset$\;
  \ForEach{$b \in \mathcal{A}_{goal},\ b\neq a$}{
    $X \leftarrow \textsc{Nodes}(R_a)\cap\textsc{Nodes}(R_b)$\;
    $CP[\textsc{Key}(R_a)] \leftarrow CP[\textsc{Key}(R_a)] \cup X$\;
    \ForEach{$v\in X$}{
      $CP\_Agents[v] \leftarrow CP\_Agents[v]\cup\{a,b\}$\;
    }
  }
}
\Return $(CP,\ CP\_Agents)$\;
\end{algorithm}

\subsection{Fragment Planner}
\label{ssec:fragment_planners}

The fragment planner is a priority-based planning approach which similarly to the PBS planner, utilises local priority orderings generated during planning rather than pre-planning global ordering as in traditional prioritised planning. 
The fragment planner, however, utilises spatially-focused resource-centric priority ordering rather than agent-agent generative orderings in PBS. Further, whilst both utilise global occupancy modelling, the fragment planner does this globally as a one-time event constructing a single global occupancy map, whereas PBS does this with repeated occupancy updates during iterative replanning. The Fragment planner then goes through a route fragmentation process where generated routes are split into fragments on contention points and spatially-local fragment ownership is awarded through priority orderings.

The core algorithm which the Fragment Planner uses is outlined in Algorithm \ref{alg:fragment_find_routes}. In this, like with the standard prioritised planner, groups are first assigned, and each active agent is iterated through. The total system occupancy is identified, as the blocking set and each agent is given their own filtered map, which excludes the nodes occupied by other agents. Once all robots have generated routes through the environment, the algorithm moves to fragment the routes based on the points of contention.

The next stage of the core algorithm is to run an all-pairs conflict detection, wherein critical points are identified as any nodes which are contained within routes of more than one agent (Algorithm \ref{alg:fragment_get_critical_points}).

\begin{algorithm}[t]
\caption{FragmentPlanner: \textsc{AssignFragments}}
\label{alg:fragment_assign_fragments_getpriorities}
\KwIn{$CP,\ CP\_Agents$, routes $\{R_a\}$, metric $\pi$}
\KwOut{Preliminary node fragments $F[a]$}
\BlankLine
$ClaimedCP \leftarrow \emptyset,\ F \leftarrow \emptyset$\;
\ForEach{agent $a$ (planner iteration order)}{
  $collective \leftarrow [\ ],\ partial \leftarrow [\ ]$\;
  \ForEach{node $v$ in route order of $R_a$}{
    \uIf{$v \notin CP[\textsc{Key}(R_a)]$}{
      append $v$ to $partial$\;
    }
    \Else{
      $\mathcal{U}_v \leftarrow CP\_Agents[v]$\;
      $h_v \leftarrow (\pi=\texttt{time})\ ?\ \texttt{time\_to\_node}\ :\ \texttt{distance\_to\_node}$\;
      $ord_v \leftarrow \textsc{GetPriorities}(\emptyset,\mathcal{U}_v,\emptyset,h_v,v)$\;
      $owner \leftarrow ord_v[0]$\;

      \uIf{$owner=a \land v\notin ClaimedCP$}{
        append $v$ to $partial$\;
        $ClaimedCP \leftarrow ClaimedCP \cup \{v\}$\;
      }
      \Else{
        \If{$partial\neq[\ ]$}{append $partial$ to $collective$}
        $partial\leftarrow[v]$;\ \textbf{break}\;
      }
    }
  }
  \If{$partial\neq[\ ]$}{append $partial$ to $collective$}
  $F[a]\leftarrow collective$\;
}
\Return $F$\;
\end{algorithm}

Once critical points are identified, the fragment planner, as outlined in Algorithm \ref{alg:fragment_assign_fragments_getpriorities} generates 
resource-centric (prioritising to resolve conflicting nodes as shared resources)
priority orderings to align conflict points to agents. This is constructed by greedily resolving each critical point to the higher priority agent, compiling fragments as it moves through the route and splitting into new fragments whenever the agent is unable to maintain continuous ownership. For this implementation, two key agent prioritisation heuristics are implemented for evaluation. The default space-only-aware\textbf{} version 
utilises shortest path distances to the critical nodes for spatial priority assignment, whilst the \textbf{space-and-time-aware} version utilises the robot's nominal speed to assign critical points to the robot based on arrival ordering.

The Fragment Planner algorithm completes with each agent holding a list of sub-paths (fragments), and each agent is allowed to execute only the first fragment. Formally, given a route $R_a = \langle v_0, \dots, v_k \rangle$, the route is decomposed into an ordered set of fragments $\{F_a^1, F_a^2, \dots, F_a^m\}$. Where a route is unattainable due to, for instance, the first node being blocked by another, the robot is moved into a passive inactive state with an empty fragment, where it will repeat planning at each planning instance until a route is available.



\section{Experiments and Results}
\label{sec:results}
The proposed fragment planners and the other MRPP approaches were evaluated in multiple simulated trials involving robotic fleets of different sizes on a topological map representation of four tunnels and mapped routes to a packhouse in a commercial strawberry farm in Kent, UK. This topological map contains over 3.6 km of traversible paths spanning 1050 unidirectional edges connecting 414 nodes. Bidirectional edges between two arbitrary nodes $n_1$ and $n_2$ are represented by two unidirectional edges, one from $n_1$ to $n_2$ and the other from $n_2$ to $n_1$. 


The experiments consisted of sets of trials of one hour duration, for a fixed robotic fleet size. The fleet size was varied from 5 to 10, resulting in 6 sets of trials over 6 hours for one MRPP algorithm. In each trial, a new task is randomly generated at one of the unoccupied nodes whenever a robot completes its navigation target. These 6 sets of trials were repeated for all the 7 multi-robot path planning approaches (Naïve Planning, three prioritised planning, PBS planning and two Fragment Planners) to evaluate the variation in performance across the agent-centric and resource-centric prioritisation approaches.


For each task, the optimal route from the assigned robot was recorded, calculated from the robot's start node to the navigation target node on an empty graph (assuming no other agents are present, giving the optimal route). The path executed by the robot throughout the task's lifetime (that depends on the MRPP approach being used) was also recorded and on task completion, the ratio of the initial optimal route and the executed route was calculated according to (\ref{eq:1}) to represent the $i^{th}$ task's path optimisation efficiency ($\text{POE}_i$). This is averaged over all tasks within the trial to get the representative $POE_{task}$ from that trial.

\begin{equation}
\mathrm{POE}_i = \frac{d^{\mathrm{exec}}_i}{d^{\mathrm{opt}}_i},
\qquad
\mathrm{POE}_{\mathrm{task}} = \frac{1}{N}\sum_{i=1}^{N} \mathrm{POE}_i
\label{eq:1}
\end{equation}

However, as the number of tasks in each trial may vary depending on the dynamic conditions, a more accurate representation of the average POE than the $\text{POE}_{\text{task}}$, is calculated as the ratio of the total optimal distance for all completed tasks and the total executed distance in (\ref{eq:2}). This value ($\text{POE}_{\text{avg}}$) was also recorded along with the total tasks executed over the test period.

\begin{equation}
\mathrm{POE_{avg}}=\frac{\sum_{i=1}^{N} d^{\mathrm{exec}}{i}}{\sum_{i=1}^{N} d^{\mathrm{opt}}_{i}}
\label{eq:2}
\end{equation}


Collisions were detected on a topological level when the robots share the same node, edge or crossing edges to ensure the MRPP algorithms work as expected. However, this detection was disabled in the tests with the Naïve planner, because in this approach, all robots navigate through their optimal routes, resulting in collision scenarios. However, it represents the best possible throughput for all robots, as they complete their tasks at the earliest.

\begin{figure}[t!]
    \centering
    \includegraphics[width=1\linewidth]{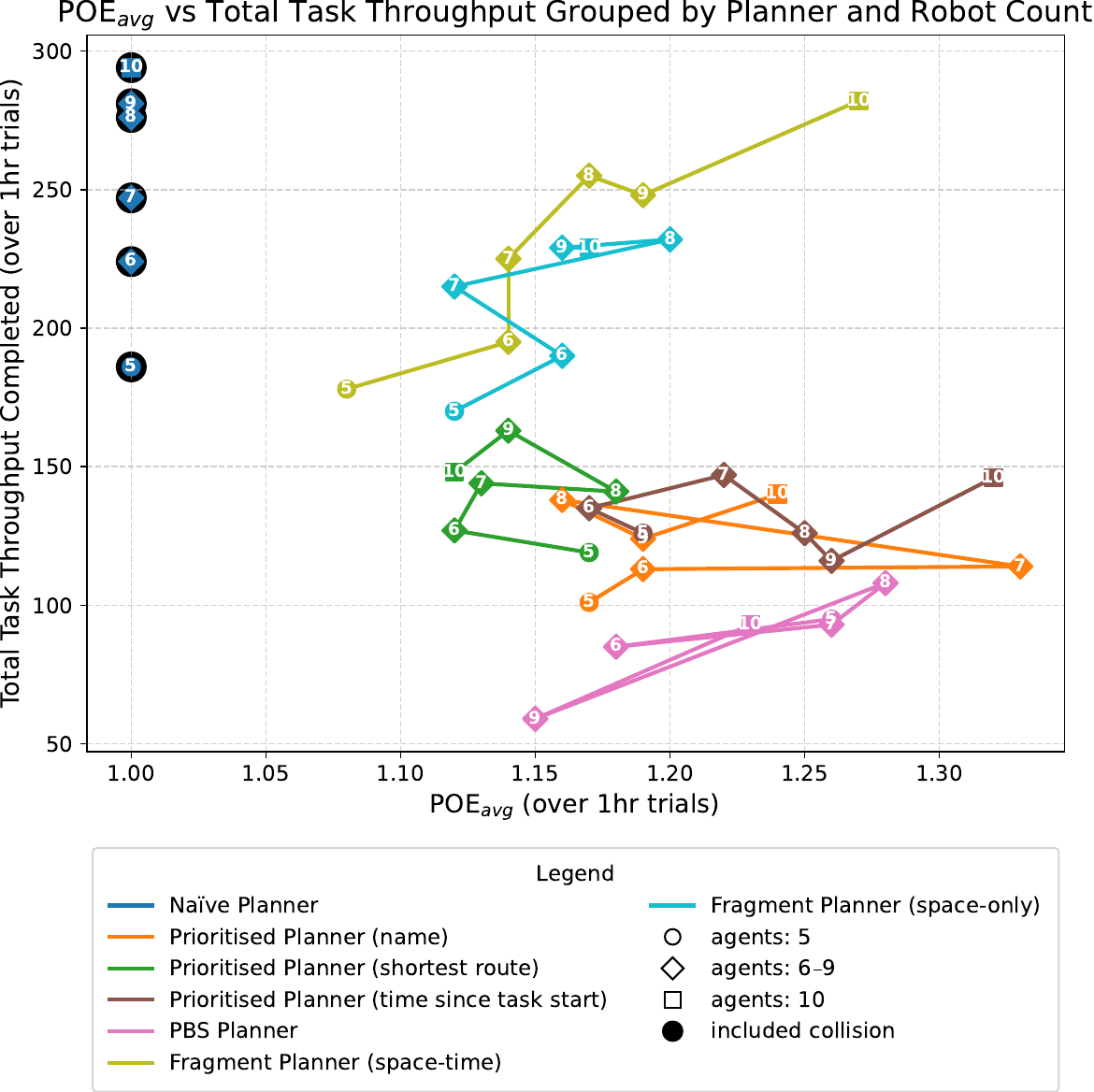}
    \caption{Total task throughput vs average path optimisation efficiency of varying multi robot path planning approaches with varying numbers of robots.}
    \label{fig:homogenous_tests}
\end{figure}

A full comparison between the $\text{POE}_{\text{avg}}$ and the total task throughput is presented for each combination of total number of robots and planner type in Figure \ref{fig:homogenous_tests}. Ideally, each additional robot would result in an increase in the total number of task executions proportional to the fleet size without reducing the efficiency of the fleet. From the results, it can be seen that both the fragment planners (FPs) configurations show positive correlations, indicating that under these planning strategies, adding more robots improves throughput with a tolerable increase in per-robot inefficiency. Conversely, the results using Prioritised Planners (PPs) show little correlation between $POE_{avg}$ and total task throughput. It can be interpreted from this, that adding more robots reduces system efficiency through worsening resource distribution.

\begin{figure}[t!]
    \centering
    \includegraphics[width=1\linewidth]{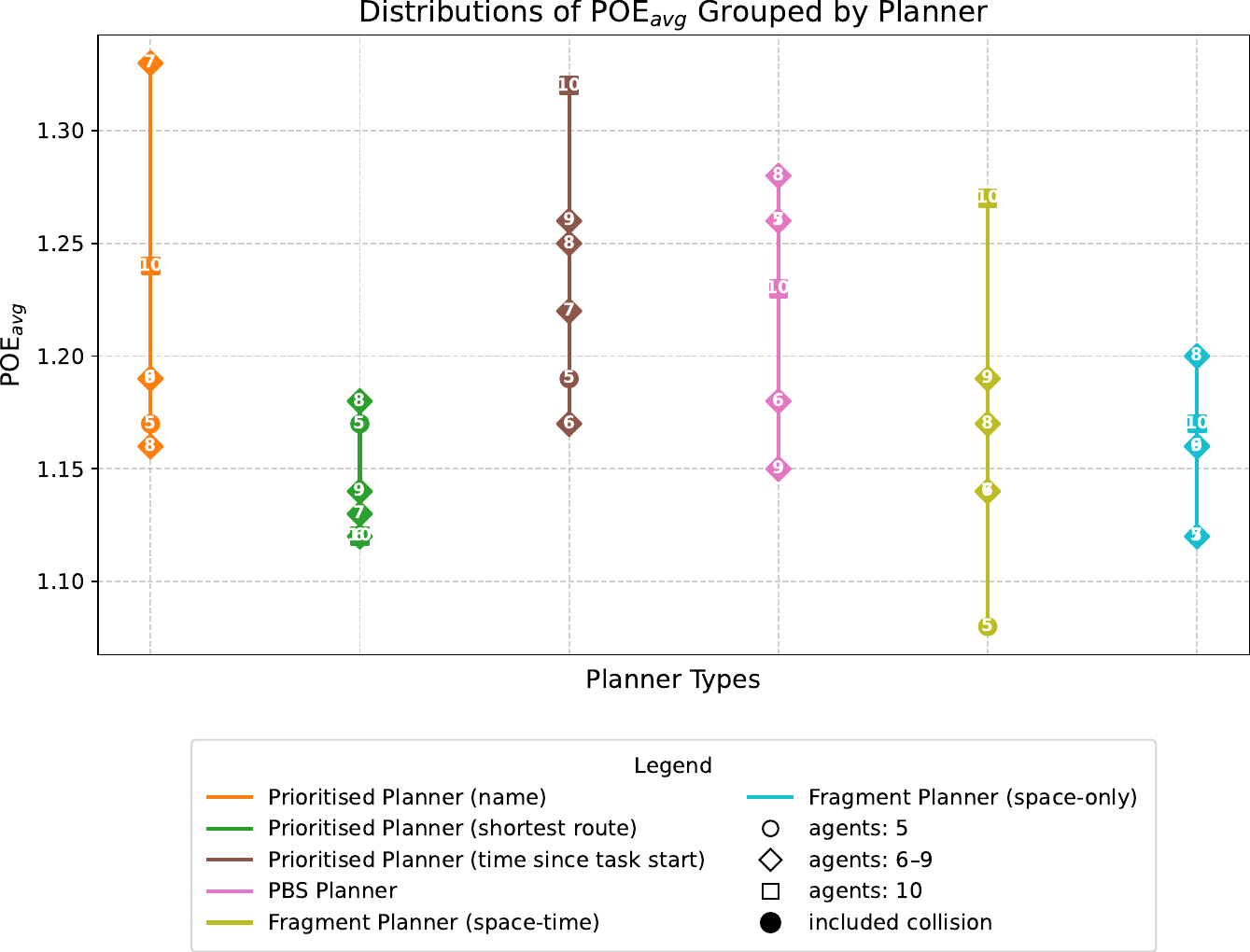}
    \caption{Distributions of average path optimisation efficiency across varying robot counts, grouped by planner.}
    \label{fig:results_homogenous_planner_type_vs_poe}
\end{figure}

In the evaluation of a coordinated system's ability to manage resources, planner efficiency is a key indicator. Figure \ref{fig:results_homogenous_planner_type_vs_poe} shows the distribution of $\text{POE}_{\text{avg}}$ values achieved per planner type over the trials. The results show FP (space-time-aware) is able to achieve the best $\text{POE}_{\text{avg}}$ at $1.08$. However, the range of FP (space-time-aware) is high compared to the more predictable distributions under PP (shortest route) and FP (space-only-aware), both achieving an average at around $1.15\pm0.05$, compared to the range of all other planners at $\pm0.2$.

\begin{figure}[t!]
    \centering
    \includegraphics[width=1\linewidth]{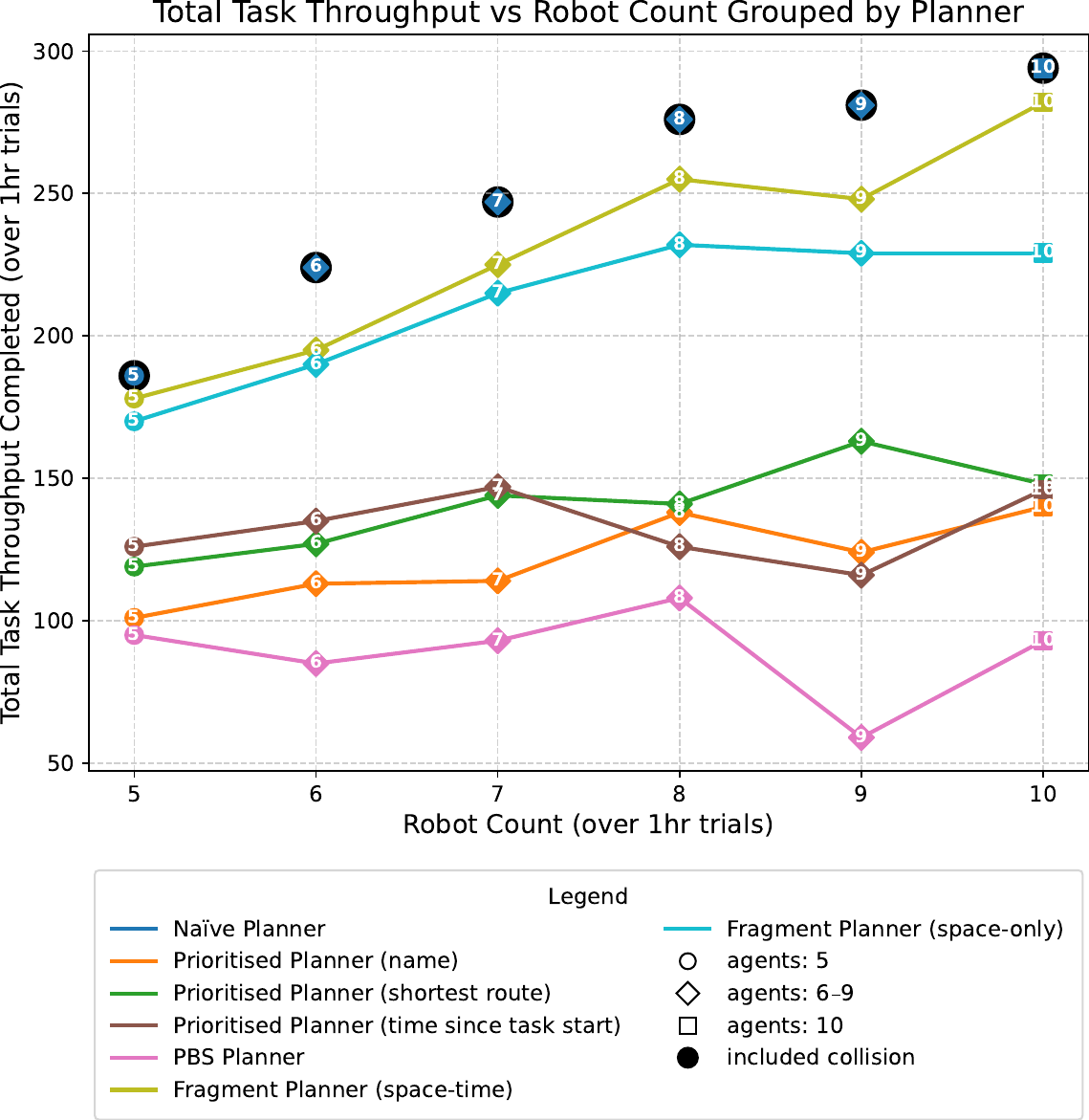}
    \caption{Total task throughput over varying robot counts grouped by planner.}
    \label{fig:results_homogenous_robot_count_vs_throughput}
\end{figure}

Whilst $\text{POE}_{\text{avg}}$ is a representative indicator of system efficiency, throughput makes a much better indicator of system efficacy. To better highlight the variation between total task throughput, Figure \ref{fig:results_homogenous_robot_count_vs_throughput} presents the variation in task completion rates according to agent counts. These results highlight that both FP implementations had significantly higher throughputs compared to the baseline PP and PBS planners. This evidences the impact of the spatial fragmentation in improving route execution, where rather than fully waiting for route availability, the FP approaches encourage partial route progression, thus reducing wait times and enabling a higher throughput. As such, FP (space-time-aware) shows the best potential throughput gains, inferring as the number of robots increase, that throughput will remain unsaturated.

The PBS planner shows the worst performance in throughput under these conditions. However as $\text{POE}_{\text{avg}}$ is stable as the number of robots increases. It is likely that the reduced throughput is result of PBS being unable to generate non-conflicting paths causing robots to wait for a route. Given that PPs generally perform better, it can thus be reasoned that priority instability from local rescheduling is playing a key role in the reduced quality. The PBS planner has been proven to have higher throughput in free space execution. So, it can be reasoned that the reduced quality be associated to the corridor-like environment causing a lack of space for interleaving.

Figure \ref{fig:results_homogenous_robot_count_vs_throughput} shows the variability in throughput per planner as the number of robots increases. For each robot count, the Naïve Planner (NP) results indicate the highest potential task throughput, if no collision may occur, even when every robot is able to immediately execute their optimal route. It is generally understood that achieving such a throughput (even with a coupled planner) is not guaranteed. However, it is an appropriate indicator for the feasible maximum throughput and can act as an appropriate baseline for evaluation.

To assess how well the planners function with respect to maximum potential efficiency, Table \ref{tab:throughput_vs_nieve} displays the relative throughput with respect to NP. The results show clearly the strengths of the FPs in achieving higher efficiency, with at 10 robots, the FP (space-time-aware) achieving over 95\% of the potential demonstrated by the NP. Contrasted with the PBS, which at 9 robots is barely able to achieve a fifth of the potential demonstrated by the NP.

\begin{table}[ht]
\centering
\caption{Throughput as \% of Naïve Planner for different fleet size}
\label{tab:throughput_vs_nieve}
\begin{tabular}{lcccccc}
\toprule
Planner & 5 & 6 & 7 & 8 & 9 & 10 \\
\midrule
PP (name) & 54.3 & 50.45 & 46.15 & 50.00 & 44.13 & 47.62 \\
PP (shortest route) & 63.98 & 56.70 & 58.30 & 51.09 & 58.01 & 50.34 \\
PP (task start time) & 67.74 & 60.27 & 59.51 & 45.65 & 41.28 & 49.66 \\
PBS & 51.08 & 37.95 & 37.65 & 39.13 & 21.00 & 31.63 \\
FP (space-only) & 91.4 & 84.82 & 87.04 & 84.06 & 81.49 & 77.89 \\
FP (space-time) & \textbf{95.7} & \textbf{87.05} & \textbf{91.09} & \textbf{92.39} & \textbf{88.26 }& \textbf{95.92} \\
\bottomrule
\end{tabular}
\end{table}



\section{Conclusion}\label{sec:conc}

When planning under lifelong conditions in corridor-like environments, it is critical that system-wide rhythm emerges. Given the density and frequency of bottlenecks in such an environment, planners utilising agent-level global priority lock-in and agent-centric priority imbalance struggle to efficiently resolve the inherently spatial contentions with agent-level evaluations, which can lead to notable reductions in system-wide efficacy. This is presented in extensive multi-robot trials highlighting how a resource-centric framing through the proposed Fragment Planners can enable a significantly higher system efficacy through route fragmentation. This is demonstrated at up to 95\% of the best-case task throughput and efficiencies as low as 8\% above optimal routes. This suggests that contentions under lifelong planning in corridor-like environments are most effectively modelled with resource-centric framing.

\section*{Acknowledgements}
This work was supported by AgriFoRwArdS CDT, under the Engineering and Physical Sciences Research Council [EP/S023917/1].

\printbibliography

\end{document}